\documentclass[10pt,journal,compsoc,twoside]{IEEEtran}
\ifCLASSOPTIONcompsoc
  \usepackage[nocompress]{cite}
\else
  \usepackage{cite}
\fi

\ifCLASSINFOpdf
  \usepackage[pdftex]{graphicx}
\else
\fi

\usepackage{amsmath}
\usepackage[ruled,vlined]{algorithm2e}

\usepackage{amssymb}
\usepackage[table,xcdraw]{xcolor}

\ifCLASSOPTIONcompsoc
  \usepackage[caption=false,font=footnotesize,labelfont=sf,textfont=sf]{subfig}
\else
  \usepackage[caption=false,font=footnotesize]{subfig}
\fi

\usepackage{multirow}   

\begin{document}
\title{Constrained Design of Deep Iris Networks}

\author{Kien~Nguyen,~\IEEEmembership{Member,~IEEE,}
        Clinton~Fookes,~\IEEEmembership{Senior Member,~IEEE,}
        Sridha~Sridharan,~\IEEEmembership{Life Senior Member,~IEEE,}        
\IEEEcompsocitemizethanks{\IEEEcompsocthanksitem K. Nguyen, C. Fookes and S. Sridharan are with Image and Video Research Laboratory, SAIVT, School of Electrical Engineering and Computer Science, Queensland University of Technology, Brisbane, QLD, 4000, Australia.\protect\\
E-mail: \{k.nguyenthanh,c.fookes,s.sridharan\}@qut.edu.au}

\thanks{Manuscript received May 23rd, 2019}}

%
%

\markboth{IEEE Transactions}%
{Nguyen \MakeLowercase{\textit{et al.}}: Constrained Design of Deep Iris Networks}
%



\IEEEtitleabstractindextext{%
\begin{abstract}
Despite the promise of recent deep neural networks in the iris recognition setting, there are vital properties of the classic IrisCode which are almost unable to be achieved with current deep iris networks: the compactness of model and the small number of computing operations (FLOPs). This paper re-models the iris network design process as a constrained optimization problem which takes model size and computation into account as learning criteria. On one hand, this allows us to fully automate the network design process to search for the best iris network confined to the computation and model compactness constraints. On the other hand, it allows us to investigate the optimality of the classic IrisCode and recent iris networks. It also allows us to learn an optimal iris network and demonstrate state-of-the-art performance with less computation and memory requirements.
\end{abstract}

}

\maketitle

\IEEEdisplaynontitleabstractindextext

%
\IEEEpeerreviewmaketitle

\section{Introduction}
\label{sec:introduction}
Deep neural networks are extremely effective at automatic feature learning for object recognition by leveraging the power of high capacity models, vast amounts of data and high-end computing infrastructure \cite{DLsurvey,DLbiometrics}. Recently, deep networks have shown to be able to automatically learn discriminative features with promising recognition accuracy in the iris recognition setting \cite{OTS_CNN_Iris,DeepIrisNet,IrisFCN, DeepIris}. However, when compared with handcrafted approaches such as the classic IrisCode \cite{IrisCodeTheory}, it is pertinent to ask a question: do we really need networks that \emph{deep} with {tens or even hundreds of layers} and {millions of parameters and floating-point operations} in the iris recognition setting? Recall that the IrisCode has been very successful with only a few parameters, \emph{e.g.} orientations and scales of Gabor filters, and it has been very cheap to run (30 ms on a CPU for iris image analysis and creation of an IrisCode) \cite{HWSWcodesign}. 



Despite its superior accuracy, current deep iris networks are unable to achieve two vital properties of the classic IrisCode: the compactness of the model and the small number of computing operations (FLOPs), which are major driving forces for iris recognition to be the favorite choice compared with other biometric modalities. This is due to the inherent large size of the deep networks with millions of parameters and hundreds of layers that are potentially required to achieve the desired accuracy. The benefits of automatic feature engineering come at the cost of significant computational power and memory requirements. Because automatic feature learning is highly desired in the iris recognition context (to both remove the pitfalls in feature design and automatically discover the best feature representation directly from the data), it is urged to understand the trade-off between the superior accuracy by automatic feature engineering and and the excessive computation and network model size we have to adopt for this benefit. 


In addition, there are a vast number of possible architectures (\emph{i.e.} number of layers, filters, connections, etc.) of a neural network, which can be millions in case of deep networks. In the iris recognition setting, the existing architectures can range from several layers (FeatNet \cite{IrisFCN}) to tens of layers (DeepIrisNet \cite{DeepIrisNet}, off-the-shelf CNNs \cite{OTS_CNN_Iris}) with millions of possible connections between them. Redundant layers, connections and filters would lead to non-compact feature representations and unnecessary computation. This raises two important questions: (i) how can we determine the optimal architecture? and (ii) how do we fully automate the network design process, \emph{i.e.} automatic feature learning and automatic architecture learning? These are important issues that must be addressed to reach the full potential of deep networks in the iris recognition setting. 


One possible solution to the above problems is to leverage recent advances in {\em neural architecture search} (NAS) to search for the network architecture that can achieve the best accuracy. Modern neural architecture search relies on reinforcement learning and evolution theory to explore the architecture space to gradually evolve the architecture to better ones. For example, Zoph \emph{et al.} designed a reinforcement learning agent to explore the architecture search space to find the optimal configurations \cite{NAS,NASNet}. Real \emph{et al.} relied on evolution theory to gradually evolve the network architecture toward higher performing ones \cite{EvolNet}. 

Existing NAS only focuses on improving the accuracy. However, there are other vital factors which are critical to the success in real-world applications. In the iris recognition setting, these include two properties mentioned above: the model compactness and the computation. Both factors have been the main driving forces to the success of modern iris recognition \cite{DaugmanInformationTheory,Daugman07,IRINA}. Taking these factors into account is important especially for practical applications. For example, if the iris recognition system is stored and processed on a high-computing platform, then there needs to be no restrictions on model compactness or matching speed, and the design process can target the best accuracy. In contrast, if the iris recognition system is performed on embedded or mobile systems with limited resources, model compactness and matching speed have to be prioritized.

There are two main objectives of this paper: (1) Develop an algorithm to search for the optimal deep iris network that both satisfies the pre-defined constraints in computation and model compactness and achieves the highest possible performance; and (2) Use this algorithm to investigate the optimality of the classic IrisCode and recent iris networks. To achieve these objectives, we re-visit existing NAS, and re-interpret it with additional constraints to model the design as a constrained optimization problem. Solving this constrained optimization problem allows us to automatically discover the optimal network (the network achieves the highest achievable accuracy within the constraints of the computation and model compactness). In addition, applying this algorithm with two constraints similar to those of the existing approaches, we are able to: (i) understand the optimality of the classic handcrafted IrisCode and recent deep iris networks; and (ii) optimize an existing network architecture to achieve the best accuracy under the same computational and memory cost.


Our major contributions can be summarized as follows:
\begin{itemize}
\item Our work is the first effort to consider automatic architecture search for deep iris recognition networks. 
\item We re-interpret the search for network architecture and parameters as a constrained optimization with design constraints related to the compactness of the model along with the matching speed. 
\item Our approach results in the full automation of deep iris network design, in both feature learning and architecture learning.
\item We discover an optimal deep network for iris recognition achieving state-of-the-art performance with less computation and memory required compared to existing deep iris networks. 
\item We provide a way to understand the optimality of the classic IrisCode approach.
\end{itemize}

The remainder of the paper is organized as follows. Section~\ref{sec:Related} discusses related feature learning and architecture learning for iris recognition; Section~\ref{sec:optimization} presents how we model the network design as a bi-level constrained optimization problem, Section~\ref{sec:Experiments} illustrates our experimental results; and the paper is concluded in Section~\ref{sec:Conclusion}.

\section{Related work}
\label{sec:Related}
This section reviews the existing attempts in the literature on applying deep learning approaches for iris recognition and architecture learning via neural architecture search.

\subsection{Feature learning for iris recognition}
A number of deep networks have been proposed to take advantage of their automatic feature engineering capability to automatically learn feature representation for iris images, aiming to improve recognition performance of the iris biometric system. Liu \emph{et al.} proposed a DeepIris network of 9 layers including one pairwise filter layer (similar to a convolutional layer), one convolutional layer, two pooling layers, two normalization layers, two local layers and one fully-connected layer \cite{DeepIris}. Gangwar \emph{et al.} employed more advanced layer types to create two DeepIrisNets for the iris recognition task \cite{DeepIrisNet}. The first network, DeepIrisNet-A, contains 8 convolutional layers (each followed by a batch normalization layer), 4 pooling layers, 3 fully connected layers and two drop-out layers. The second network, DeepIrisNet-B, adds two inception layers to increase the modeling capability. 
To deal with the lack of a large-scale dataset, transfer learning can be used. Nguyen \emph{et al.} experimented a wide range of off-the-shelf CNNs, which had won the ImageNet challenges (AlexNet, VGG, Inception, ResNet, DenseNet), on the iris recognition task \cite{OTS_CNN_Iris}. They showed that these off-the-shelf CNN features can be successfully transferred to the iris recognition task, achieving state-of-the-art recognition accuracy. In contrast, Zhao \emph{et al.} argued that CNNs may be not optimal for the iris recognition task since iris texture, which is believed to be random, does not exhibit structural information or meaningful hierarchies \cite{IrisFCN}. Hence they proposed a fully convolutional network named FeatNet without any fully connected layers to retain spatial correspondence with the original input image, leading to higher recognition rates. 

All these deep iris networks have their architectures handcrafted, which means potential redundant connections, filters and layers, and no best accuracy guarantee. Learning both a feature representation and an optimal architecture is challenging considering the huge number of potential architectures and their parameters. This challenge will be addressed in this paper. 

\begin{figure*}
\centering
\includegraphics[width=1.7\columnwidth]{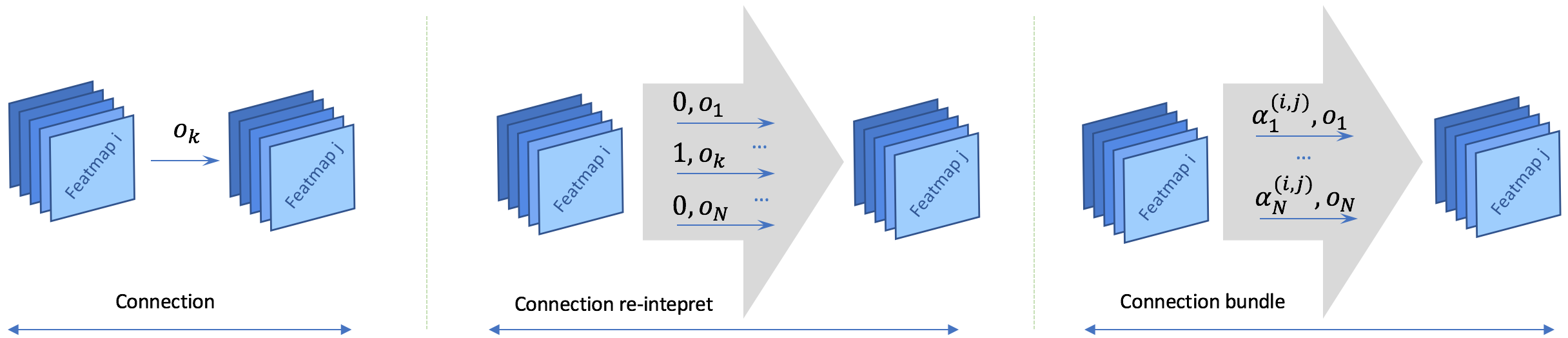}
\caption{Re-interpreting layer connections. Traditionally, two layers, $i$ and $j$, are connected by one operation selected from the operation set, $\mathbb{O} = \{o_1,...,o_N\}$. This means activating only operation, $o_k$, in the operation set with a coefficient $1$, while disabling all others with a coefficients $0$. The coefficients only take a discrete value $\{0,1\}$. We relax this to allow the coefficients $\alpha_k^{(i,j)}$ to take a continuous value between $[0,1]$. This re-interpretation enables us to make the architecture search space continuous. }
\label{fig:connectionbundle}
\end{figure*}

\subsection{Neural Architecture Search}
There does not exist any work in network architecture search in the iris recognition setting. In general deep learning, modern architecture search approaches usually rely on Evolution Theory and Reinforcement Learning (RL) to design searching policies. For example, Real \emph{et al.} brought in the ideas from the natural evolution process to gradually update the architecture of the network to achieve higher accuracies \cite{EvolNet,AmoebaNet}. At each iteration, a number of architectures, called population, are investigated, the best architecture is mutated (\emph{i.e.} randomly add/remove layers) to generate a new child architecture to be added into the population. Subsequently, the worst architecture \cite{EvolNet} or the oldest architecture \cite{AmoebaNet} is discarded to generate a new population. This algorithm is considered as a discrete optimization process. In contrast, Zoph \emph{et al.} trained a RNN controller to iteratively sample candidate architectures, and trains them to convergence to measure their performance on the desired task. The controller then uses the performance as a guiding signal to find more promising architectures \cite{NAS,NASNet}. Parameter sharing can be forced on all child models to improve the search speed at a slight cost of performance \cite{ENAS}. Both evolution and RL approaches are very computationally expensive despite their remarkable performance. For example, obtaining a state-of-the-art architecture for CIFAR-10 required 1800 GPU days of RL \cite{NASNet} or 3150 GPU days of evolution \cite{AmoebaNet}. 

One of our main motivations arises from a recent body of works on gradient-based architecture search \cite{gradientNAS,darts} where the problem is encoded as a bi-level optimization problem with one level to optimize based on the architecture and the other level to optimize based on the weights of the chosen architecture. This strategy is able to discover high-performing architectures achieving high classification accuracy in only tens of GPU hours. This strategy is highly desired in the iris recognition scenario to search for the highest-accuracy architecture. However, it is not directly applicable to the iris recognition scenario because in contrast to the natural image recognition scenario, the approaches in the iris recognition scenario need to take into account design constraints as discussed earlier to be applicable in real-world applications. In this paper, we introduce a new design procedure to address these challenges.

It is also worth noting that two related tasks: hyperparameter optimization and network simplification/pruning are simpler tasks considering the dimension of the parameters compared with the whole architecture. In addition, a recent body of works on resources-aware network design such as MorphNet \cite{MorphNet} and NetAdapt \cite{NetAdapt} have considered resource constraints in the design process. However, these techniques work on simplifying a pre-trained model to match the resource constraints, which is much simpler than an architecture search from scratch as being addressed in this paper.  





\section{Designing constraints}
\label{sec:optimization}

A network architecture, $\alpha$, is defined as a directed acyclic graph consisting of an ordered sequence of $L$ nodes. Each node, $x_j$, in the graph is a latent representation, a.k.a a feature map in the network. The first node is the input node, which is the input image. The final node is the output. Each intermediate node is computed as a summary of network operations applied on its predecessors,
\begin{equation}
\label{equ:nodesummary}
x_j = \underset{i<j}{\sum} o^{(i,j)}(x_i),
\end{equation}
where $o^{(i,j)}$ is a candidate network operation with value taken from an operation set $\mathbb{O}$. The operation set can include a convolutional operator, $conv$; a pooling operator, $pool$; a skip connection operator, $Identity$; and a no connection indicator, $zero$. Each convolutional operator is followed by a batch norm operation by default. The size of the convolutional kernel can vary, \emph{e.g.} $3\times3$ and $3\times5$. There are two types of convolution: traditional convolution and its dilated version \cite{DilatedConv}. The $Identity$ operation will function as a skip connection similar to the ResNet architecture \cite{ResNet}, which adds the original signal from the input feature map to the output feature map. The $Zero$ operation will model the lack of connection between two feature maps. 

With this notation, the network design is interpreted as two tasks: (1) searching for a set of network operations $\{o^{(i,j)} : \; j:1..L \;\; \text{and} \;\; i:1..j \;\; \}$; and (2) searching for the operation weights (a.k.a parameter values) to achieve the highest performance network in the iris recognition setting.

For each network, we consider two constraints:
\begin{itemize}
\item \textbf{Compactness of model:} number of parameters: $P$, which is directly related to memory required: $M_P$,
\item \textbf{Computation:} number of FLOPS: $K$.
\end{itemize}
Our aim is to design a network that achieves the best accuracy conditioned on these constraints. 


\subsection{Re-interpreting layer connections}

Considering $o^{(i,j)}$ is discrete, as shown in the literature, searching in the discrete space is extremely computational heavy and may result in missing the optimal point \cite{gradientNAS,continuosNAS}. We employ one adjustment to make the search space continuous by assigning a coefficient $\alpha_o^{(i,j)}$ for each candidate operations in the operation set $\mathbb{O}$. Other than activating a single operation while all others are disabled between two nodes $(i,j)$, we activate all candidate operations in the operation set but only one operation is strongly encouraged with a high coefficient value while others are strongly discouraged with small coefficient values. We call this a connection bundle as shown in Figure~\ref{fig:connectionbundle}. In the traditional discrete case, $\alpha_o^{(i,j)}$ takes only one of two values, $\{0,1\}$. The most popular choice is one of $\alpha_o^{(i,j)}$ is $1$ while all others are $0$. In our continuous case, $\alpha_o^{(i,j)}$ can take any value in the range of $[0,1]$. This adjustment has been employed in the gradient-based architecture learning approaches \cite{darts}.

The categorical summary in Equation~\ref{equ:nodesummary} can now be interpreted as a softmax over all possible operations,
\begin{equation}
\overline o^{(i,j)}(x) = \sum_{o \in \mathbb{O}} \frac{exp(\alpha_o^{(i,j)})}{Z} o(x),
\end{equation}
where the denominator $Z$ is the normalization factor defined as $Z = \sum_{o' \in \mathbb{O}} exp(\alpha_{o'}^{(i,j)})$. After this relaxation, the task of architecture search reduces to learning a set of continuous variables $\alpha = \{ \alpha^{(i,j)} \}$. At the end of the search, a discrete architecture can be obtained by replacing the pseudo operation $\overline o^{(i,j)}$ with the strongest operation, \emph{i.e.} $o^{(i,j)} = \text{argmax}_{o \in \mathbb{O}} \;\; \alpha_o^{(i,j)}$. 

\begin{figure*}
\centering
\includegraphics[width=1.8\columnwidth]{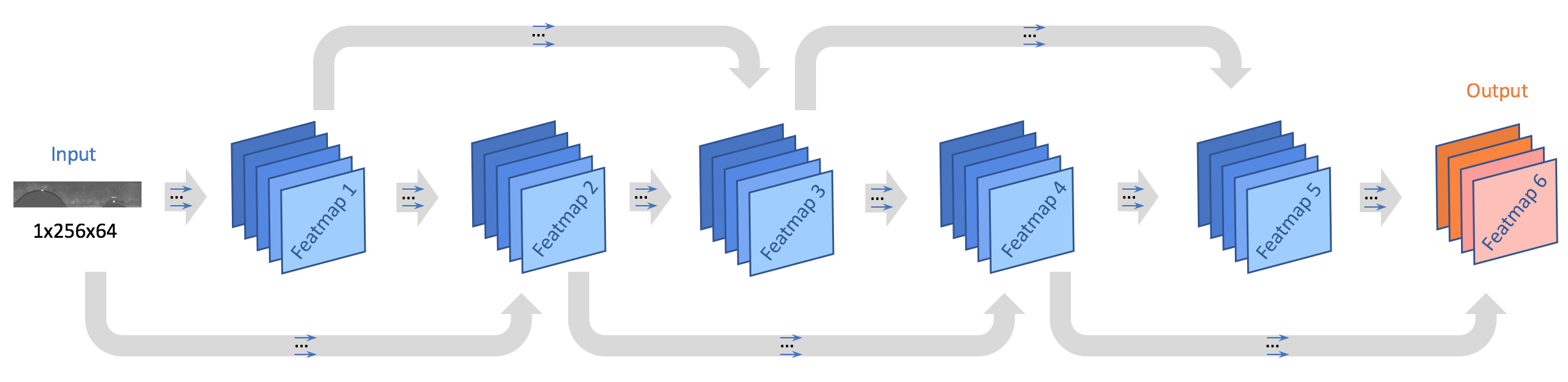}
\caption{One search architecture example: with connection bundles, the architecture search becomes an optimization task to find the best set $\{\alpha_o^{(i,j)}\},i,j=1..L,i<j$ to minimize the loss.}
\label{fig:searcharchitecture}
\end{figure*}

One example of an architecture search space is illustrated in Figure~\ref{fig:searcharchitecture}.

\subsection{Network design as a constrained optimization}
Denote $L_{train}$ and $L_{val}$ as the training and validation losses respectively. Both losses are determined by not only the architecture, $\alpha$, but also the weights, $w$, of the network. The goal of our network design is to find $\alpha^*$ that minimizes the validation loss $L_{val}(w^*,\alpha^*)$, where the weight, $w^*$, associated with the architecture are obtained by minimizing the training loss, $w^*(\alpha) = \text{argmax}_w L_{train}(w,\alpha^*)$.   

Now the network design is a double learning problem. Taking the constraints mentioned before, the learning process can be represented as a bi-level constrained optimization problem as follows:

\begin{equation}
\underset{\alpha}{\text{min}}  \;\; L_{val}(w^*(\alpha),\alpha) + |\alpha|,
\end{equation}
\begin{eqnarray}
\textrm{subject to: } &\#&FLOP(\alpha) < K, \\
\textrm{and } &\#&Parameter(\alpha) < P, \\
\textrm{and } &w^*(\alpha)& = \underset{w}{\text{argmin}} \label{equ:loweroptimization}\;\; L_{train}(w,\alpha),
\end{eqnarray}
where the validation data is used to search for the architecture and the training data is used to search for the weights of an architecture. This is a two-level optimization problem where: (1) the lower level optimization problem is: searching for the best weights of the existing architecture to minimize the training loss; and (2) the upper level optimization problem is: searching for the best network architecture that, with its best weights, minimizes the validation loss. 

This bi-level optimization has also arisen in hyperparameter optimization in the network design such as in \cite{bilevelHO} but in simple forms where the dimensions of the scalar-valued hyperparameters are substantially smaller than the architecture dimension. 

\subsection{Iterative solution}
Solving the bilevel optimization exactly is prohibitive as it would require recomputing $w^*$ by solving the lower problem (\ref{equ:loweroptimization}) whenever there is a change in $\alpha$. An iterative gradient-based solution has been shown to be effective \cite{BilevelOptimization}. However, popular gradient-based approaches such as Stochastic Gradient Descent (SGD) \cite{ConvexOptimization} is not applicable here due to the presence of inequality design constraints. To solve this, we propose an algorithm to solve this constrained bi-level optimization problem based on Projected Gradient Descent (PGD) \cite{PGD} as in Algorithm~\ref{alg:ConstrainedBilevel}. The gradient of the loss in the train subset, $\nabla_w L_{train}(w,\alpha)$, is used to update the weights, $w$, to learn the best weight for the current architecture. The gradient of the loss in the validation subset, $\nabla_{\alpha} L_{val}(w-\xi \nabla_w L_{train}(w,\alpha),\alpha)$, is used to guide the architecture update, where $\xi$ is a coefficient. The updated architecture, $\alpha$, is tested on the computation and memory constraints. If it does not satisfy the constraints, $\alpha$ will be projected into the constraint domain to generate a new architecture that satisfies the constraints. This iterative gradient-based approach has been applied in \cite{darts} but without constraints.

\begin{algorithm}
\caption{Constrained deep iris network design}
\SetAlgoLined
\SetKwInOut{Input}{Inputs}
\Input{Constraints: \newline
       - maximum number of FLOPs: K \newline
       - maximum number of parameters: P}
\SetKwInOut{Output}{Outputs}
\Output{+ optimal architecture $\alpha^*$ \newline 
        + its weights $w^*$ \newline
        + optimal performance metrics: $EER^*$ }
 Create a mixed operation $\overline o^{(i,j)}$ parameterized by $\alpha_o^{(i,j)}$ for each edge (i,j)\;
 \While{not converged}{
  instructions\;
  1. Update weights $w$ by descending $\nabla_w L_{train}(w,\alpha)$\;
  2. Update architecture $\alpha$ by descending $\nabla_{\alpha} L_{val}(w-\xi \nabla_w L_{train}(w,\alpha),\alpha)$\;
  3. If $\alpha$ does not satisfy the \#FLOPs and \#parameters constraints, project it to the constraint domain to generate a new architecture
 }
 Replace $\overline o^{(i,j)}$ with $o^{(i,j)} = \text{argmax}_{o \in  \mathbb{O}} \; \alpha_o^{(i,j)}$ for each edge (i,j)
\label{alg:ConstrainedBilevel}
\end{algorithm}

\section{Experimental results}
\label{sec:Experiments}
We have conducted our experiments on three publicly available datasets: 
\begin{itemize}
\item \underline{ND-CrossSensor-Iris-2013 dataset \footnote{https://sites.google.com/a/nd.edu/public-cvrl/data-sets}:} is the largest public iris dataset in the literature in terms of the number of images \cite{NDIRIS0405}. It contains 116,564 iris images captured by the LG2200 iris camera from 676 subjects.
\item \underline{CASIA-Iris-Thousand dataset \footnote{http://biometrics.idealtest.org}:} contains 20,000 iris images from 1,000 subjects, which were collected using the IKEMB-100 camera from IrisKing \cite{web:CASIA}. 
\item \underline{UBIRIS.v2 iris dataset \footnote{http://iris.di.ubi.pt/ubiris2.html}:} contains 11,102 iris images from 261 subjects with 10 images each subject. The images were captured under unconstrained conditions (at-a-distance, on-the-move and on the visible wavelength), with realistic noise factors \cite{UBIRISv2}.
\end{itemize}

\begin{figure}
\centering
\includegraphics[width=\columnwidth]{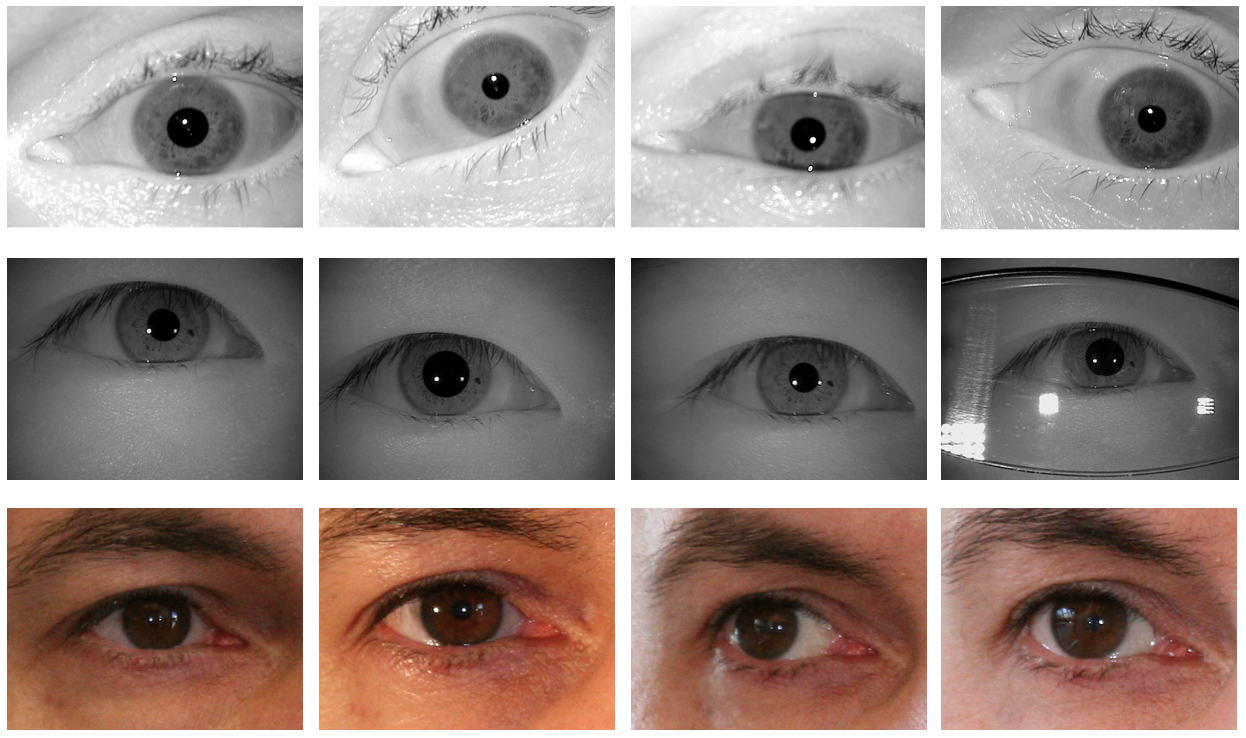}
\caption{Sample images from ND-CrossSensor-2013 (1st row), CASIA-Iris-Thousand (2nd row) and UBIRIS datasets (3rd row).}
\label{fig:samples}
\end{figure}

\begin{table*}
\caption{Statistics of three datasets, ND-CrossSensor-2013, CASIA-Iris-Thousand and UBIRIS.v2 in this research.}
\centering
\label{tab:datasets}
\begin{tabular}{|l|l|l|l|l|l|l|l|} 
\hline
                    & \textbf{\# Subjects} & \textbf{\# Images} & \textbf{Distance}  & \begin{tabular}[c]{@{}l@{}}\textbf{Image}\\\textbf{Resolution}\end{tabular}& \begin{tabular}[c]{@{}l@{}}\textbf{Iris}~\\\textbf{Diameter}\end{tabular}& \textbf{Wavelength} & \begin{tabular}[c]{@{}l@{}}\textbf{Subject}~\\\textbf{Cooperation}\end{tabular}  \\ 
\hline
ND-CrossSensor-2013 & 676         & 111,564   & Close-up   & 640x480& 200~& NIR& Highly                                                                                                                                                                \\ 
\hline
CASIA-Iris-Thousand & 1,000       & 20,000    & Close-up   & 640x480& 180& NIR& Highly                                                                                                                                                              \\ 
\hline
UBIRIS.v2           & 261         & 11,102    & 4-8 meters & 800x600& 180-80& Visible& Less                                                                                                                                                       \\
\hline
\end{tabular}
\end{table*}

Some examples and statistics of three datasets are presented in Figure~\ref{fig:samples} and Table~\ref{tab:datasets}.

Three experiments will be performed for validation. First, the proposed algorithm will be employed to investigate the optimality of the existing approaches. By targeting the computation and/or model size of the existing approaches as the constraints for the search, our constrained search algorithm automatically discovers the network that achieves the highest achievable accuracy. Comparing this accuracy with the accuracy of the existing approaches, their optimality can be validated or rejected. The classic IrisCode and a recent deep iris network will be sequentially used for demonstration in the first two experiments presented in Section~\ref{sec:case1} and ~\ref{sec:case2}. The third experiment is to search for a network with competitive accuracy to the state-of-the-art approaches with less computation and model size requirements. 


\subsection{Experimental setup} 
\label{sec:preprocessing}
We first pre-process the iris images by segmentation and normalization. The iris image is first segmented using two circles of the inner and outer boundaries of the iris region, detected by an integro-differential operator as,
\begin{equation}
max_{r,x_{0},y_{0}}\big|G_{\sigma}(r)*\frac{\partial}{\partial r}\oint _{r,x_{0},y_{0}} \frac{I(x,y)}{2\pi r}ds \big|,
\end{equation}
where $I(x,y)$ and $G_{\sigma}$ denote the input image and a Gaussian with a standard deviation $\sigma$, respectively. The symbol $*$ denotes a convolution operation and $r$ represents the radius of the circular arc $ds$, centered at the location $(x_0,y_0)$. This operation detects circular edges by iteratively searching for the maximum responses of a contour defined by the parameters $(x_0, y_0, r)$. We subsequently normalize the segmented iris regions to a fixed size by a rubber-sheet model \cite{Daugman03}. This process is carried out by re-mapping the iris region, $I(x,y)$, from the raw Cartesian coordinates $(x,y)$ to the dimensionless polar coordinates $(r,\theta)$ as, $I(x(r,\theta),y(r,\theta)) \rightarrow I(r,\theta)$, where $r$ is in the unit interval [0,1], and $\theta$ is an angle in the range of [0,2$ \pi $]. $x(r,\theta)$ and $y(r,\theta)$ are defined as the linear combination of both pupillary $(x_p(\theta),y_p(\theta))$ and limbic boundary points $(x_s(\theta),y_s(\theta))$ as,
\begin{eqnarray}
x(r,\theta)=(1-r) x_p(\theta)+r x_s(\theta),\\
y(r,\theta)=(1-r) y_p(\theta)+r y_s(\theta).
\end{eqnarray}
This normalization step also helps to reduce the rotations of the eye (\emph{e.g.}, due to the head movement), to simple translation during matching. The corresponding noise mask is also normalized to facilitate matching in the later stages. For this segmentation and normalization, we used an open-source software, USIT v2.2, from the University of Salzburg \cite{USIT2} to segment, normalize and generate a normalized images with a fixed size of $64\times512$ pixels.


\begin{figure*}
\centering
\includegraphics[width=1.6\columnwidth]{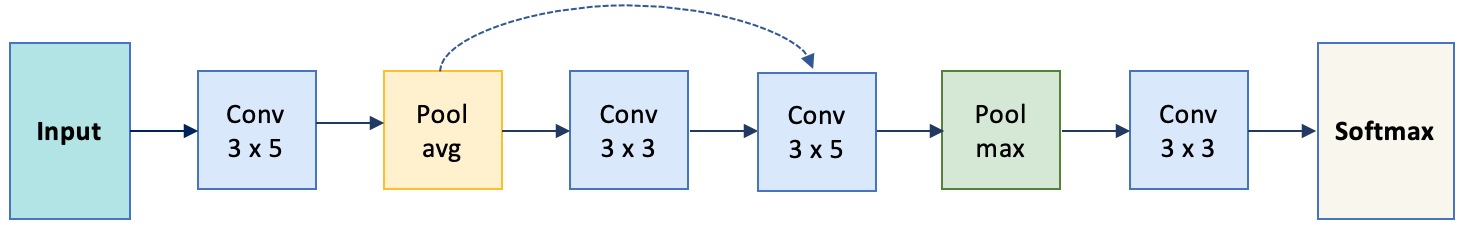}
\caption{The architecture that achieves the same accuracy level with the handcrafted IrisCode. Notice it achieves the same performance level with 8 times additional computation.}
\label{fig:irisnet}
\end{figure*}

\subsection{Performance metrics}
To report the performance, we rely on False Rejection Rate (FRR) and Equal Error Rate (EER). In this work, FRRs at False Acceptance Rate (FAR) = 0.1\% are experimented and reported due to its popular adoption in the field. 

State-of-the-art iris networks employ two types of losses: (1) cross-entropy loss \cite{DeepIrisNet,OTS_CNN_Iris} and (2) pairwise loss \cite{DeepIris,IrisFCN}. 
\begin{itemize}
    \item Cross-entropy loss: the important property of the cross-entropy loss is using the same identities in the training and testing datasets, which means sample-disjoint but not subject-disjoint. This is shown in the softmax classifier of \cite{DeepIrisNet} and the SVM classifier of \cite{OTS_CNN_Iris}. Hence to be comparable with the state of the art, we first divide our dataset into the sample-disjoint but not subject-disjoint training and testing subsets.
    \item Pairwise loss: the pairwise loss measures the similarity or dissimilarity between two input images, deciding whether they are from the same class or not. This loss allows us to have unseen subjects, \emph{i.e.} not present in the training phase, in the test phase. This loss has been shown to be effective for iris recognition \cite{DeepIris,IrisFCN}. Hence we also split the data into subject-disjoint training and testing subsets.
\end{itemize}

In summary, there are two data splitting schemes in this work: (1) sample-disjoint splitting and (2) subject-disjoint splitting. Depending on the task to be performed, our experiments will apply one of these two schemes. For the sample-disjoint scheme, we split images of each subject into 70\% of the images for training, 10\% of the images for validation and 20\% of the images for testing. For the subject-disjoint scheme, we split 70\% of the subjects for training, 10\% of the subjects for validation and 20\% of the subjects for testing.

\textbf{Intra-dataset performance}
We perform the intra-dataset experiment on the ND-CrossSensor-Iris-2013 dataset, not the other two datasets, due to its large size suitable for training. The training subset is to train a network architecture to find the best weights. The validation subset is used to find the best network architecture. The testing subset is used to report the intra-dataset performance. 

The classic IrisCode is used as a handcrafted baseline. Our implementation achieves a $FRR = 3.76\%$ at $FAR = 0.1\%$ and an $EER=1.75\%$ on the ND-CrossSensor-2013 dataset, which is comparable to the state of the art implementation \cite{IrisFCN}.

\textbf{Cross-dataset performance}
The best network learned is further investigated for generalization capability through training in one dataset and testing on others. The best network discovered in the ND-CrossSensor-Iris-2013 dataset is tested on the other two datasets, CASIA-Iris-Thousand and UBIRIS.v2 to understand its generalizability. We do not perform network search on the CASIA and UBIRIS datasets as their small number of images will restrict the search space.

\subsection{Case 1: Handcrafted - IrisCode}
\label{sec:case1}

\begin{figure*}
\centering
\includegraphics[width=1.7\columnwidth]{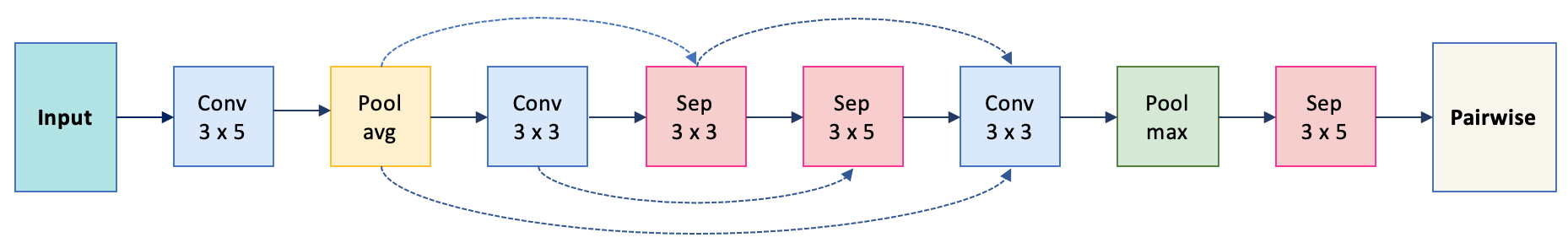}
\caption{The optimal network architecture that is discovered by our constrained search algorithm.}
\label{fig:case3}
\end{figure*}

Firstly, we are interested to see how well the deep networks perform if they have to limit their computation akin to the one in the classic handcrafted IrisCode \cite{Daugman07}. We impose one computation constraint, \emph{i.e.} the maximum number of FLOPs, to be akin to the one in the classic IrisCode, and investigate the best accuracy a deep network could achieve compared to the accuracy of the IrisCode. We run the constrained design algorithm to find the best network architecture yielding the highest accuracy or smallest EER conditioned on the IrisCode computation. 

\textbf{Operation set}
We apply popular operations, which are widely used in the existing deep iris networks, in the operation set $\mathbb{O}$: $3\times3$ and $3\times5$ convolution, $2\times2$ max pooling, $2\times2$ average pooling, $Identity$ and $Zero$. All operations are of stride 1.

\textbf{Loss choice}
We apply the most popular softmax classifier with a classification cross-entropy loss. A majority of previous deep iris recognition networks use this classification loss trained over a set of known iris identities and then take the intermediate bottleneck layer as a representation beyond the set of identities used in training. We use the classification probabilities as scores on known subjects in the test set. Varying the score threshold generates different operating points in the ROC curve. To be succinct, we only present the EER in the first two experiments. The ROC curve will be presented in the third experiment when comparing with the state-of-the-art approaches.  

\textbf{Dataset} The experiment is performed on the ND-CrossSensor-Iris-2013 dataset. Due to the cross-entropy loss, the sample-disjoint scheme is applied to split the dataset into  70\% of the images in each subject for training, 10\% of the images in each subject for validation and 20\% of the images in each subject for testing.

\begin{table}[h]
\centering
\caption{Case 1 constrained design task.}
\begin{tabular}{|l|l|l|}
\hline
\textbf{Inputs}                                                       & \textbf{\begin{tabular}[c]{@{}l@{}}Algorithm\\ hyperparameters\end{tabular}}                                                                  & \textbf{Outputs}                                              \\ \hline
\begin{tabular}[c]{@{}l@{}}K = $K_{IrisCode}$\\ P = $Inf$\end{tabular} & \begin{tabular}[c]{@{}l@{}}- $\mathbb{O}$ = \{$3\times3$, $3\times5$ $conv$; \\ $2\times2$ $max$ pool; $2\times2$ $avg$\\ pool; $Identity$; $Zero$\}\\ - Loss = cross-entropy\end{tabular} & \begin{tabular}[c]{@{}l@{}}- $\alpha^*$\\ - $w^*$\\- $EER*$\end{tabular} \\ \hline
\end{tabular}
\label{tab:case1}
\end{table}

The constrained design task with input constraints, hyperparameters on the operation set and the loss choice, and the outputs are summarized in Table~\ref{tab:case1}. We found that the best network constrained by the number of FLOPs of the classic IrisCode achieves an $EER = 16.7$, which is 10 times the $EER$ achieved by the IrisCode. Two interesting points can be inferred here: (i) deep networks struggle when the computation is strictly limited. Fundamentally, strictly limited computation is detrimental to the modeling capacity and learning algorithms; and (ii) this reinforces the effectiveness of the classic IrisCode in terms of both computation and recognition accuracy.


We subsequently pose a question related to how much additional computation and memory we have to sacrifice to achieve similar or even better accuracies than the handcrafted counterpart. We gradually increased the constraint $K$ values by a $K_{IrisCode}$ step and the initial number of layers, then reported the EERs as shown in Table~\ref{lbl:computation}. The same level of accuracy is only achieved with $8$ times the additional computation amount with the network architecture discovered in Figure~\ref{fig:irisnet}. This aligns with the well-known universal approximation theory of deep networks \cite{IrisUniversality}, which states that a deep network can approximate any function given enough resources. This illustrates the main characteristic of these deep networks: automatic feature engineering comes at the cost of heavy computation and memory requirements.


\begin{table}
\centering
\caption{Computation vs. EER of deep architectures vs. IrisCode. The first row presents the computation ratio between the deep network and IrisCode $K_R=\frac{K_{deeparchitecture}}{K_{IrisCode}}$; the second row presents the best $EER$ achieved by under the selected computation constraints; and the third row presents the accuracy ratio between the deep network and IrisCode $EER_R=\frac{EER_{deeparchitecture}}{EER_{IrisCode}}$.}
\begin{tabular}{|l|l|l|l|l|l|l|l|} 
\hline
$K_R$     & 2.0    & 3.0    & 4.0   & 5.0  & 6.0  & 7.0 & 8.0 \\ \hline
EER      & 16.2   & 15.1   & 13.3  & 8.9  & 5.6  & 2.9 & 1.4 \\ \hline
$EER_R$  & 9.3    & 8.6    & 7.6   & 5.1  & 3.2  & 1.7 & 0.8\\
\hline
\end{tabular}
\label{lbl:computation}
\end{table}


\subsection{Case 2: Deep learning - ResNet18} 
\label{sec:case2}
Secondly, we are interested to see how effective the state-of-the-art deep networks perform in terms of architecture design. In other words, under the same computation and memory with the existing state-of-the-art networks, what is the best accuracy we can achieve for the iris recognition task. Nguyen \emph{et al.} analyzed layer by layer performance of the landmark networks which have won the ImageNet challenge since 2012 to the iris recognition tasks \cite{OTS_CNN_Iris}. Despite being pre-trained on the ImageNet, they have shown competitive performance on the iris recognition task with transfer learning. We choose one of the landmark networks, called ResNet18, for experiments due to its simplicity and uniformity in the layer connections and wide adoption in the field \cite{ResNet}. We apply the proposed constrained design algorithm to search for the network with the best accuracy that can be yielded bounded by the computation, $K_{ResNet18}$, and the number of parameters, $P_{ResNet18}$, of the ResNet18 network. We employ the same operation set, the cross-entropy loss and the data splitting scheme as in Case 1. The constrained design task is summarized in Table~\ref{tab:case2}. Our algorithm discovers a new architecture that achieves higher accuracy than the original ResNet, $EER = 1.12\%$ vs. $1.29\%$ and $FRR = 2.23\%$ vs. $2.58\%$ with the same level of computation and memory.

\begin{table}[h]
\centering
\caption{Case 2 constrained design task.}
\begin{tabular}{|l|l|l|}
\hline
\textbf{Inputs}                                                       & \textbf{\begin{tabular}[c]{@{}l@{}}Algorithm\\ hyperparameters\end{tabular}}                                                                  & \textbf{Outputs}                                              \\ \hline
\begin{tabular}[c]{@{}l@{}}K = $K_{ResNet18}$\\ P = $P_{ResNet18}$\end{tabular} & \begin{tabular}[c]{@{}l@{}}- $\mathbb{O}$ = \{$3\times3$, $3\times5$ $conv$; \\ $2\times2$ $max$ pool; $2\times2$ \\$avg$ pool; $Identity$; $Zero$\}\\ - Loss = cross-entropy\end{tabular} & \begin{tabular}[c]{@{}l@{}}- $\alpha^*$\\ - $w^*$\\- $EER*$\end{tabular} \\ \hline
\end{tabular}
\label{tab:case2}
\end{table}

\subsection{Case 3: State-of-the-art}
\label{sec:case3}
We also want to see whether we can achieve competitive accuracy compared with the state-of-the-art deep iris networks.   

\textbf{Operation set}
We apply popular operations in the operation set $\mathbb{O}$: $3\times3$ and $3\times5$ convolution, $3\times3$ and $3\times5$ dilated convolution, $2\times2$ max pooling, $2\times2$ average pooling, $Identity$ and $Zero$. All operations are of stride 1.

\textbf{Loss choice}
We leverage the most recent advance in the loss design for biometrics by using a pairwise loss called Extended Triplet loss as investigated in \cite{FaceNet,IrisFCN}. Compared with classification losses, pairwise losses \emph{directly reflect what we want to achieve, \emph{i.e.} to train the representation to correspond to iris similarity}. This results in irises of the same person having small distances and irises of different people having larger distances.

\textbf{Dataset} The experiment is performed on the ND-CrossSensor-Iris-2013 dataset. Due to the pairwise loss, the subject-disjoint scheme is applied to split the dataset into  70\% of the subjects for training, 10\% of the subjects for validation and 20\% of the subjects for testing.


The constrained design task is summarized in Table~\ref{tab:case3}. Running on a single Nvidia GTX 1080Ti GPU, our algorithm discovered a network as presented in Figure~\ref{fig:case3} in 26 hours. We compare the discovered network with the state-of-the-art approaches in four metrics: two in accuracy ($FRR$ and $EER$), one computation ($K$) and one memory ($P$).

\begin{table}[h]
\centering
\caption{Case 3 constrained design task.}
\begin{tabular}{|l|l|l|}
\hline
\textbf{Inputs}                                                       & \textbf{\begin{tabular}[c]{@{}l@{}}Algorithm\\ hyperparameters\end{tabular}}                                                                  & \textbf{Outputs}                                              \\ \hline
\begin{tabular}[c]{@{}l@{}}K = $K_{FeatNet}$\\ P = $P_{FeatNet}$\end{tabular} & \begin{tabular}[c]{@{}l@{}}- $\mathbb{O}$ = \{$3\times3$, $3\times5$ $conv$; \\ $3\times3$, $3\times5$ $dilated$ conv; \\ $2\times2$ $max$ pool; $2\times2$ \\$avg$ pool; $Identity$; $Zero$\}\\ - Loss = pairwise \end{tabular} & \begin{tabular}[c]{@{}l@{}}- $\alpha^*$\\ - $w^*$\\- $EER*$\end{tabular} \\ \hline
\end{tabular}
\label{tab:case3}
\end{table}

\begin{table}[]
\centering
\caption{Comparison with state of the art approaches on ND-CrossSensor-2013 dataset.}
\begin{tabular}{|l|c|c|l|l|}
\hline
                                    & \textbf{FRR}    & \textbf{EER}    & \textbf{K}  & \textbf{P}  \\ \hline
IrisCode           & 3.76\%          & 1.75\%          &  0.5M       & 5     \\ \hline
\multicolumn{5}{|l|}{}  \\[-2ex] \hline
DeepIris \cite{DeepIris}            & 2.62\%          & 1.31\%          &  20M        & 192K  \\ \hline
FeatNet \cite{IrisFCN}              & 1.79\%          & 0.99\%          &  30M        & 13K    \\ \hline
\textbf{Ours}                       & \textbf{1.32\%} & \textbf{0.68\%} &  19M        & 11K  \\ \hline
\end{tabular}
\label{tab:intra-datasetSOTA}
\end{table}

\begin{figure}[!t] 
\centering
\includegraphics[width=\columnwidth]{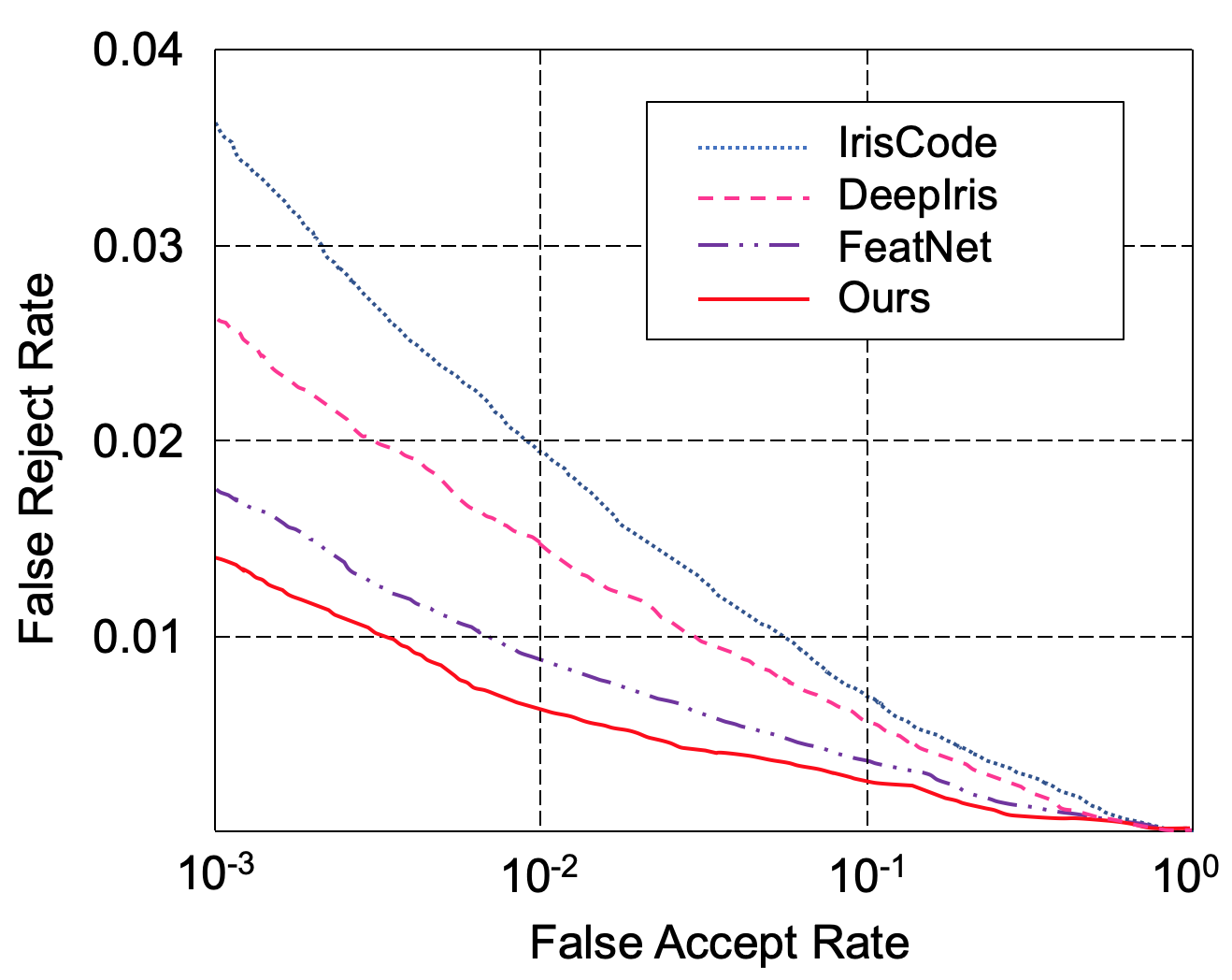}
\caption{ROC curves for comparison with other deep learning feature representations on the test set of the ND-CrossSensor-2013 dataset. \emph{Best viewed in color.}} 
\label{fig:comparisonDL}
\end{figure}

We compare with two state-of-the-art deep iris networks using pairwise loss including: DeepIris \cite{DeepIris} and FeatNet \cite{IrisFCN}. Since the original models in their papers are not publicly available, we carefully implemented and trained the networks according to all the details in \cite{DeepIris,IrisFCN}. A notable difference is we use the same segmentation method from USIT v2.0 for all approaches to be comparable. The normalized iris images, with a size of $64\times512$, are fed directly to \cite{IrisFCN} since they use the same input size. The normalized iris images are resized to the expected size of \cite{DeepIris,IrisFCN} to be compatible with their network designs. The performance achieved is comparable to those reported in the papers. Table~\ref{tab:intra-datasetSOTA} and the ROC curve in Figure~\ref{fig:comparisonDL} show the discovered network outperforms all existing deep iris networks in terms of accuracy with less number of parameters and computation required.

\subsection{Generalizability of the model}
Finally, we want to understand the generalization capability of the network architecture discovered by performing a cross-dataset experiment on two smaller-size datasets, CASIA \cite{web:CASIA} and UBIRIS \cite{UBIRISv2}. While the CASIA dataset captured the NIR iris images using a different camera, the UBIRIS dataset captured the iris images with a visible camera. This demonstrates the wide range of imaging conditions to test the generalizability. Two datasets CASIA and UBIRIS are splitted into 20\% for training and 80\% for testing. Three networks, \emph{i.e.}: ours, DeepIris and Featnet, trained as per Section~\ref{sec:case3} are further fine-tuned using the training subset and tested in the testing subset of the two datasets for cross-dataset performance investigation.

The performance is presented in Table~\ref{tab:CrossDataset}. The network discovered by our architecture search algorithm outperforms the state-of-the-art approaches in both CASIA and UBIRIS datasets, illustrating a high level of generalization across different sensors, different imaging distances and different levels of subject cooperation.

\begin{table}[]
\centering
\caption{Cross-dataset performance of the network discovered.}
\label{tab:CrossDataset}
\begin{tabular}{|l|c|c|c|c|}
\hline
                        & \multicolumn{2}{c|}{\textbf{CASIA-Iris-1K}} & \multicolumn{2}{c|}{\textbf{UBIRISv2}} \\ \hline
                        & \textbf{FRR}            & \textbf{EER}            & \textbf{FRR}       & \textbf{EER}      \\ \hline
IrisCode                & 5.53\%                  & 3.46\%                  & 14.31\%            & 8.33\%            \\ \hline
\multicolumn{5}{|l|}{}                   \\ [-2ex] \hline
DeepIris \cite{DeepIris}               & 4.25\%                  & 2.17\%                  & 13.20\%            & 7.12\%            \\ \hline
FeatNet \cite{IrisFCN}                & 3.98\%                  & 1.93\%                  & 13.93\%            & 6.69\%            \\ \hline
\textbf{Ours} & \textbf{3.07\%}         & \textbf{1.54\%}         & \textbf{11.12\%}   & \textbf{5.98\%}   \\ \hline
\end{tabular}
\end{table}


\section{Conclusions}
\label{sec:Conclusion}
This paper proposes an algorithm to design a deep iris recognition network with attention to computation and memory constraints. By modeling the design process as a bi-level constrained optimization approach, our algorithm is able to search for the optimal network which achieves the best possible performance conditioned on the pre-defined computation and model compactness constraints. This algorithm enables us to investigate the effectiveness of the classic handcrafted IrisCode compared with deep network counterparts. It also enables us to further improve the existing deep iris recognition networks to achieve similar or better accuracy with the same level of computation and memory cost. The design algorithm also discovers an optimal network with competitive performance with less computation and memory required than the state-of-the-art approaches, in both intra-dataset and cross-dataset experiments. More importantly, this algorithm simultaneously achieves both automatic feature engineering and network architecture engineering, opening us to full automation in deep iris recognition network design.

{\small
\bibliographystyle{ieee}
\bibliography{irisDL}
}

%

\begin{IEEEbiographynophoto}{Kien Nguyen} is a Research Fellow at Queensland University of Technology (QUT, Australia). He has been conducting research in iris recognition for 10 years, having published in the top conferences and journals such as CVPR, PR, and CVIU. His research interests are applications of computer vision and deep learning techniques for biometrics, surveillance and scene understanding. He has been serving as an Associate Editor of the journal IEEE Access in the area of biometrics since 2016.
\end{IEEEbiographynophoto}

\begin{IEEEbiographynophoto}{Clinton Fookes} is a Professor in Vision and Signal Processing at the Queensland University of Technology. He holds a BEng (Aerospace/Avionics), an MBA, and a PhD in computer vision. He is a Senior Member of the IEEE, an AIPS Young Tall Poppy, an Australian Museum Eureka Prize winner, and a Senior Fulbright Scholar.
\end{IEEEbiographynophoto}

\begin{IEEEbiographynophoto}{Sridha Sridharan} obtained his MSc degree from the University of Manchester, UK and his PhD degree from University of New South Wales, Australia. He is currently a Professor at Queensland University of Technology (QUT) where he leads the research program in Speech, Audio, Image and Video Technologies (SAIVT).
\end{IEEEbiographynophoto}




\end{document}